\documentclass[conference]{IEEEtran}
\IEEEoverridecommandlockouts

\usepackage{graphicx}
\usepackage{booktabs}
\usepackage{amsmath,amssymb}
\usepackage{cite}
\usepackage{url}
\usepackage{algorithm}
\usepackage{algorithmic}
\usepackage{placeins}
\usepackage{float}
\usepackage[table]{xcolor}
\usepackage{array}
\usepackage{multirow}
\usepackage{eso-pic}
\usepackage[font=footnotesize,skip=2pt]{caption}

\usepackage[hidelinks]{hyperref}

\usepackage[font=footnotesize,skip=2pt]{caption}

\IEEEoverridecommandlockouts

\makeatletter
\def\ps@IEEEtitlepagestyle{%
    \def\@oddhead{%
        \hbox to \textwidth{%
            \hfill
            \parbox[t]{0.70\textwidth}{%
                \raggedleft
                \scriptsize
                12th IEEE International Conference on Sustainable Technologies
                and Engineering (IEEE iCOSTE 2026)\\
                1--3 November 2026, Cox's Bazar, Bangladesh
            }%
        }%
    }%
    \def\@evenhead{\@oddhead}%
    \def\@oddfoot{%
        \hbox to \textwidth{%
            \parbox[b]{\columnwidth}{%
                \raggedright
                \scriptsize
                979-8-3195-2804-9/26/\$31.00
                \textcopyright~2026 IEEE
            }%
            \hfill
        }%
    }%
    \def\@evenfoot{\@oddfoot}%
}
\makeatother

\vspace{0.5em}
\begin{document}

\title{An Efficient Approach to Beetle Damage Detection of Banana Using Deep Learning}
\vspace{0.5cm}
\author{
\IEEEauthorblockN{
Md. Mehedi Hasan Naeem,
Ishtiyak Ahmmad Araf,
Md. Arefin Haque Mahir,
Tushar Kanti Saha
}

\IEEEauthorblockA{
\textit{Department of Computer Science and Engineering}\\
\textit{Jatiya Kabi Kazi Nazrul Islam University}\\
Trishal, Mymensingh-2224, Bangladesh\\
mehedinaeem00@gmail.com,
ishtiyakahmmad1149@gmail.com\\
arefinmahir2026@gmail.com,
tushar@jkkniu.edu.bd
}
}

\maketitle

\begin{abstract}
Visible surface damage of banana fruits by Basilepta subcostata (the banana fruit scarring beetle) can reduce their market value. Here, manual inspection of Banana fruits is subjective and difficult to scale. This study presents a lightweight framework based on YOLOv8n deep learning model for locating individual bananas and classifying their visible surface condition as healthy or unhealthy. The term beetle damage in the paper title is used operationally for the locally observed scar-like surface condition; the model does not identify a specific beetle species or biological cause. A custom dataset of 2,473 images containing 5,827 annotated banana instances was prepared from local-market imagery in Bangladesh. The reported experiment used an 80/20 protocol with 1,982 training images and 491 test images. A pretrained YOLOv8n model was trained at $640\times640$ resolution for 50 epochs with a batch size of 4. On the test partition, it achieved 81.36\% precision, 74.30\% recall, 77.67\% F1-score, 87.04\% mAP@0.50 and 59.33\% mAP@0.50:0.95. A YOLOv8s comparison gave slightly lower F1 and mAP while using a substantially larger checkpoint. The YOLOv8n checkpoint occupied 5.96 MB and required 6.54 ms/image for inference on an NVIDIA Tesla T4. The results support YOLOv8n as a compact baseline for instance-level banana surface inspection within the studied acquisition conditions.
\end{abstract}


\section{Introduction}
\label{sec:introduction}

Banana quality is strongly influenced by external appearance, which affects
consumer preference, market value and postharvest acceptance. Surface scars,
rough regions, dark marks and related abnormalities can reduce commercial
quality. These characteristics are commonly assessed through manual visual
inspection, but such assessment can become time-consuming and inconsistent
when large numbers of fruits must be examined before sending them to the market. Recent advances in computer vision and deep learning have enabled automated
banana-quality assessment. Real-time deep-learning systems have been developed
for banana freshness grading \cite{chen2026freshness}, while CNN-based methods
have also been investigated for export-grade classification
\cite{yasin2026export}. However, image-level classification is less suitable
when several bananas appear in one image because each fruit may have a
different surface condition. Object detection can address this limitation by
localizing and classifying individual banana instances.

In Bangladesh, bananas with scarred, rough or dark-marked surfaces are
sometimes locally described as \emph{Bitol} caused by Basilepta
subcostata (the banana fruit scarring beetle). In this study, the term refers
only to visible surface condition and not to a confirmed biological diagnosis.
Banana pests, diseases, mechanical damage, and environmental factors can produce
similar visual symptoms \cite{liao2025bananaai}. Field research in Bangladesh
has reported that banana leaf- and fruit-beetle feeding can produce visible
scars and reduce fruit quality \cite{zahan2001yield}. Moreover, taxonomic
studies have identified multiple banana-scarring beetle species
\cite{prathapan2019species}. Therefore, the present work uses two operational
visual classes, \emph{healthy} and \emph{unhealthy}, without attempting species-level pest identification. Here an efficient classifier can be useful to detect healthy and unhealthy bananas. We can devise this efficient classifier if we can perform object localization effectively. 

YOLO-family \cite{sapkota2025yoloevolution} (YOLOv5, YOLOv8, YOLOv9, and so on) are popular deep learning models widely used in agricultural computer vision because
they combine object localization and classification within an efficient
one-stage framework \cite{khan2025object}. Banana appearance-quality studies
have also demonstrated the suitability of YOLO-based methods for localized
surface assessment \cite{mao2023yolobanana}. For this experiment, YOLOv8n was selected mainly because it is compact while still providing instance-level localization and classification. The banana images were not visually uniform: illumination, ripening stage, shadows, occlusion, fruit overlap and background complexity changed from scene to scene. In some cases, dark but healthy skin could resemble damaged areas, while subtle surface abnormalities were harder to localize. Comparable difficulties have been reported in natural-scene fruit detection and surface-defect analysis \cite{wu2024dneyolo,feng2023orange}. These conditions require a lightweight instance detector suitable for the classification and localization task, provided that useful localization performance could be retained at practical computational cost.

This study investigates close-range, instance-level detection of visible
banana surface damage using a locally collected and manually annotated
dataset. The main contributions are:

\begin{enumerate}
    \item A custom banana dataset with instance-level bounding-box annotations
    for healthy and visibly abnormal surface conditions called unhealthy.

    \item A lightweight deep learning framework using YOLOv8n for simultaneous banana localization
    and healthy/unhealthy surface-condition classification.

    \item Quantitative and qualitative evaluation using precision, recall,
    F1-score, mean Average Precision (mAP), diagnostic curves, confusion analysis, and computational
    efficiency.
\end{enumerate}

\section{Related Work}

Deep learning has increasingly been applied to banana quality, disease and pest analysis. Selvaraj \emph{et al.} demonstrated transfer learning based detection of banana disease and pest symptoms from field images \cite{selvaraj2019ai}. Mora \emph{et al.} later used deep learning with high resolution UAV imagery of banana Xanthomonas with detection in complex landscapes \cite{mora2024pixels}.

For postharvest quality, Sajitha \emph{et al.} modeled banana quality characteristics with a knowledge-embedded graph convolutional network \cite{sajitha2023kegcnn}, while illumination-focused work showed that lighting variation can materially affect banana-ripeness classifiers . Image segmentation has also been investigated for early banana disease identification, emphasizing the value of localized visual evidence . These studies establish the feasibility of automated banana assessment, but most target ripeness, disease symptoms or plant-level conditions rather than close-range fruit-surface localization.

Recent agricultural studies have used YOLO models for several related detection problems. In Bangladesh, YOLOv8 has been applied to multi-crop leaf-disease detection \cite{abid2024bangladesh}, while YOLOv8/YOLOv9 have been used for localized disease-severity estimation . Tang \emph{et al.} incorporated hybrid attention into YOLOv8 for fruit-ripeness detection under changing outdoor illumination and shadows .

Other YOLOv8 variants have targeted dragon-fruit disease localization , lightweight agricultural pest detection  and subtle runner-bean leaf disease . Although the crops and symptoms differ, these studies are relevant because their targets vary in scale, texture and scene background.

Fruit-detection studies also show where errors can arise in practice. YOLO-CFruit addressed fruit detection in complex natural scenes \cite{luo2024cfruit}, while DNE-YOLO studied apple detection across diverse environmental conditions and emphasized illumination, occlusion and background variation \cite{wu2024dneyolo}.

Embedded fruit-detection studies further show that accuracy must be considered together with model size and runtime \cite{karacaoglu2025kiwi}. For surface-defect inspection specifically, orange-defect detection has shown that defect scale and visual similarity can reduce separability \cite{feng2023orange}, and YOLO-Banana targeted appearance-quality grading with localized banana defects \cite{mao2023yolobanana}. Edge-oriented mango detection also demonstrates the relevance of efficient YOLO deployment in agriculture \cite{gu2024mango}.

The literature therefore supports both automated banana assessment and lightweight agricultural detection, but close-range instance detection of locally observed banana surface damage remains comparatively less explored. Table~\ref{tab:rw} highlights the distinction between representative banana studies and the present task. The current work focuses on individual fruit instances using healthy/unhealthy visual labels without claiming species-level diagnosis.

\begin{table}[H]
\centering
\caption{Representative banana studies and the present scope.}
\label{tab:rw}
\scriptsize
\setlength{\tabcolsep}{3pt}
\begin{tabular}{p{0.19\columnwidth}p{0.34\columnwidth}p{0.37\columnwidth}}
\toprule
\textbf{Study} & \textbf{Primary task} & \textbf{Difference from this work}\\
\midrule

Selvaraj \emph{et al.} \cite{selvaraj2019ai}
& Banana disease/pest detection
& Multiple plant parts; not focused on fruit-surface scarring\\

Mao \emph{et al.} \cite{mao2023yolobanana}
& Banana appearance-quality grading
& Different dataset and YOLOv5-based design\\

Mora \emph{et al.} \cite{mora2024pixels}
& Xanthomonas wilt detection
& UAV, plant-level disease assessment\\

This work
& Healthy/unhealthy fruit detection
& Close-range, instance-level surface-condition detection\\

\bottomrule
\end{tabular}
\end{table}

\FloatBarrier

\section{Materials and Methods}
\label{sec:methods}

\begin{figure}[H]
    \centering
    \includegraphics[
        width=0.85\columnwidth,
        height=0.7\columnwidth
    ]{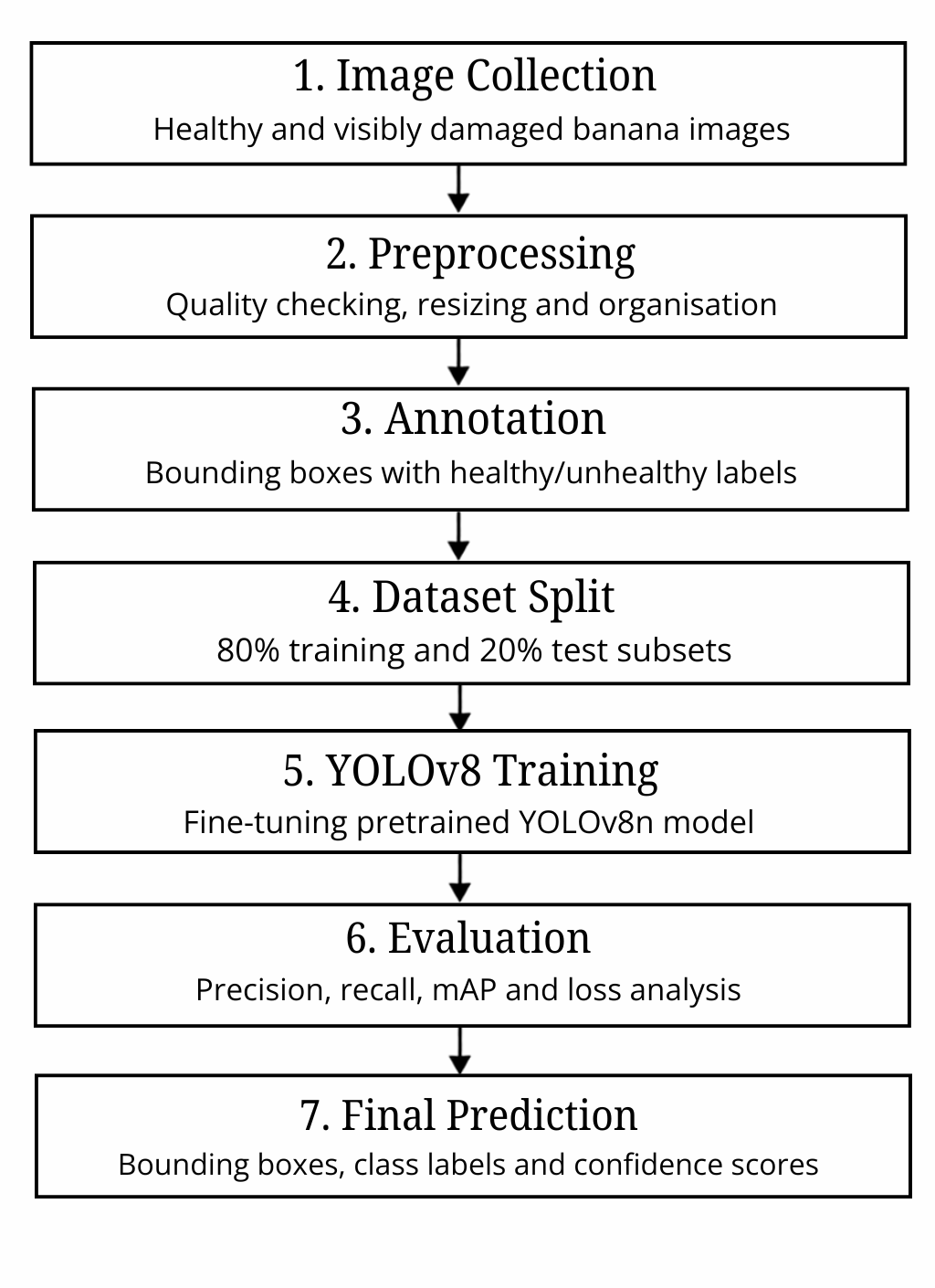}
    \caption{Overall research workflow for banana surface-condition detection.}
    \label{fig:workflow}
\end{figure}

The methodology comprised image acquisition, screening, instance annotation,
quality checks, model training and post-training evaluation on the test
partition. Fig.~\ref{fig:workflow} summarizes the procedure.

\subsection{Dataset Acquisition, Annotation and Verification}
RGB images were collected from local markets in Trishal, Mymensingh, Netrakona
and Narsingdi, Bangladesh. The collection included variation in orientation,
fruit size, ripening appearance, illumination, background and the number of
visible bananas. Severely blurred, corrupted or ambiguous images were removed
before annotation. Bounding boxes were prepared in YOLO format for two
operational classes. \emph{Healthy} denotes a visibly normal fruit surface;
ordinary ripening color alone was not intended to define the unhealthy class.
\emph{Unhealthy} requires a visible external abnormality such as a scar, rough
region, dark mark or related surface damage. These labels describe appearance
only and do not establish biological cause. The available project records do
not contain an inter-annotator agreement measurement, so no agreement statistic
is reported. Annotations followed the normalized YOLO representation \cite{jocher2023yolov8}
\begin{equation}
\mathbf{b}=(c,x_c,y_c,w,h),
\label{eq:yolo}
\end{equation}
where $c$ is the class identifier and the remaining terms define the normalized
box center and size. For an image of width $W$ and height $H$,
\clearpage
\begin{equation}
 x_c=\frac{x_{\min}+x_{\max}}{2W},\quad
 y_c=\frac{y_{\min}+y_{\max}}{2H},
\label{eq:center}
\end{equation}
\begin{equation}
 w=\frac{x_{\max}-x_{\min}}{W},\quad
 h=\frac{y_{\max}-y_{\min}}{H}.
\label{eq:size}
\end{equation}

The dataset contained 2,473 images and 5,827 annotated instances. The training
partition contained 1,982 images and 4,382 instances, while the test partition
contained 491 images and 1,445 instances. Healthy and unhealthy instance totals
were 3,233 and 2,594, respectively. Image/label pairing, malformed rows,
invalid classes and box coordinates were checked during dataset preparation.
The revised repository also records split membership and image hashes. A later
hash audit of the released files identified repeated image content across the
train/test folders; therefore, the reported metrics should be interpreted as
performance on the supplied 80/20 split rather than as a fully group-independent
estimate. This issue is retained transparently in the repository audit records.

\begin{table}[H]
\centering
\caption{Dataset distribution used in the reported experiment.}
\label{tab:data}
\footnotesize
\setlength{\tabcolsep}{4pt}
\begin{tabular}{lrrr}
\toprule
\textbf{Partition} & \textbf{Images} & \textbf{Healthy} & \textbf{Unhealthy}\\
\midrule
Train & 1,982 & 2,355 & 2,027\\
Test & 491 & 878 & 567\\
\midrule
Total & 2,473 & 3,233 & 2,594\\
\bottomrule
\end{tabular}
\end{table}

\begin{figure}[H]
\centering
\includegraphics[width=\columnwidth]{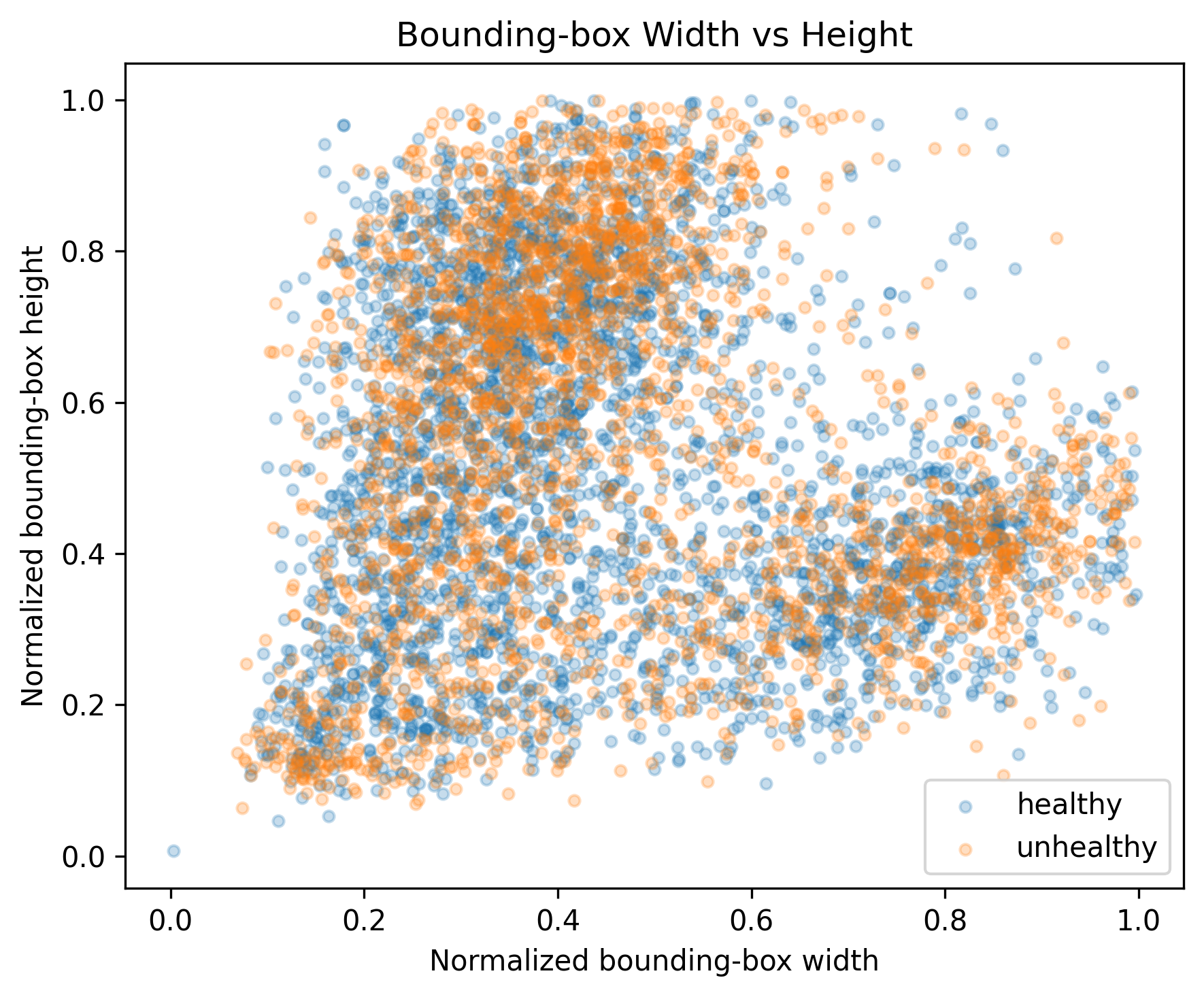}
\caption{Distribution of normalized bounding-box width and height for healthy and unhealthy banana instances.}
\label{fig:bbox_geometry}
\end{figure}

The annotation geometry was examined to characterize object scale.
Fig.~\ref{fig:bbox_geometry} shows substantial overlap between healthy and
unhealthy box geometry, indicating that size alone does not separate the
classes.

\begin{figure}[H]
\centering
\includegraphics[width=\columnwidth]{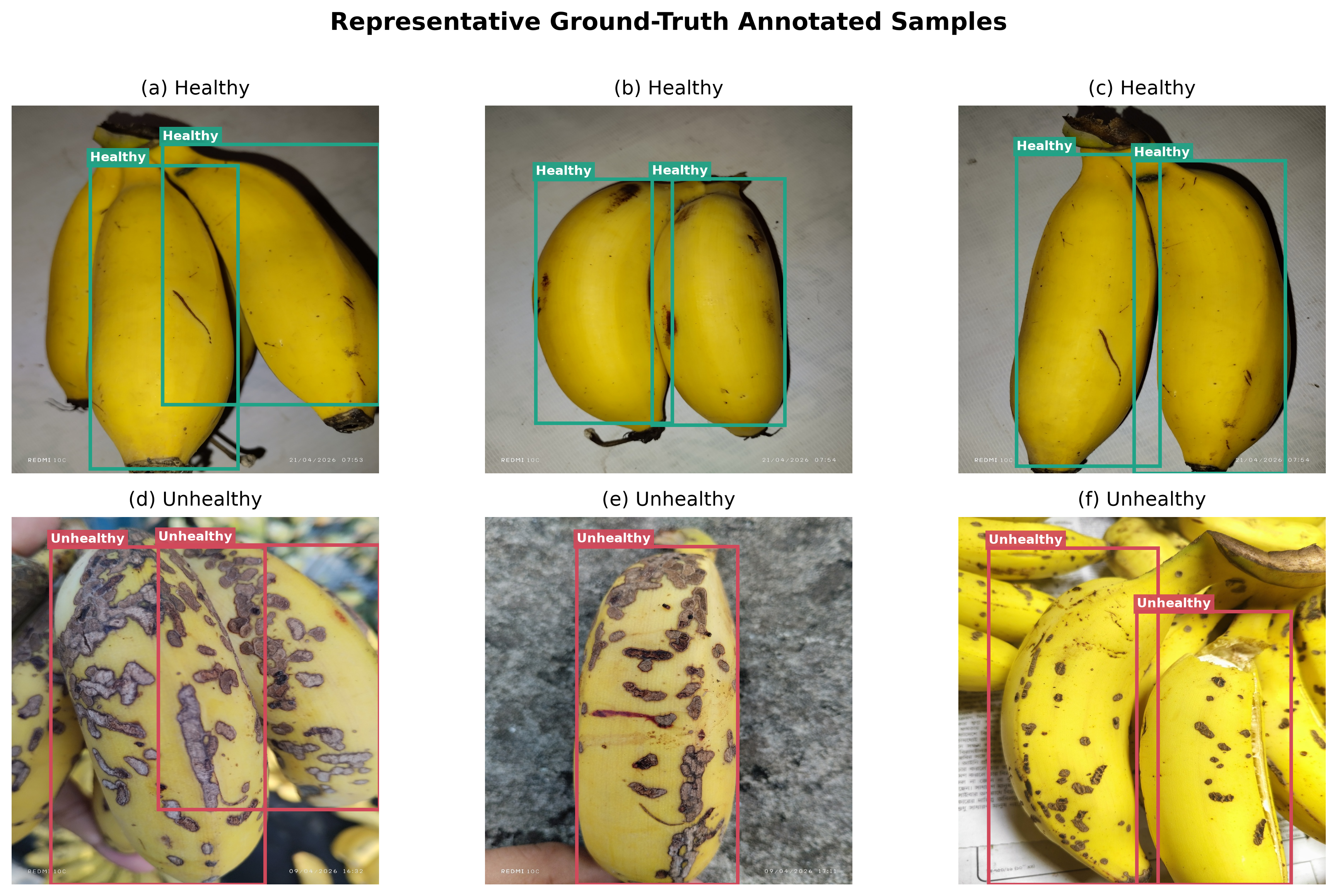}
\caption{Representative training annotations for healthy and unhealthy banana instances.}
\label{fig:ann}
\end{figure}

Representative annotated samples are shown in Fig.~\ref{fig:ann}. Multiple
bananas in the same scene can receive separate boxes and different labels.

\subsection{YOLOv8 Framework and Training}
The primary detector used pretrained YOLOv8n weights and transfer learning.
YOLOv8 combines convolutional feature extraction, C2f blocks, SPPF contextual
aggregation, multi-scale feature fusion and an anchor-free detection head
\cite{jocher2023yolov8}. The detector jointly optimizes localization and
classification:
\begin{equation}
\mathcal{L}=\lambda_b\mathcal{L}_{box}+\lambda_c\mathcal{L}_{cls}+\lambda_d\mathcal{L}_{DFL},
\label{eq:loss}
\end{equation}
where $\lambda_b$, $\lambda_c$ and $\lambda_d$ weight the bounding-box,
classification and Distribution Focal Loss components, respectively. DFL
models box locations as distributions and was introduced in the Generalized
Focal Loss framework \cite{li2020gfl}.

The reported YOLOv8n run used Ultralytics 8.4.52, pretrained weights,
$640\times640$ input, batch size 4 and 50 epochs. The saved configuration used
\texttt{optimizer=auto}, $\mathrm{lr0}=0.01$, $\mathrm{lrf}=0.01$,
momentum 0.937, weight decay 0.0005 and three warm-up epochs. Active
augmentations included HSV variation, translation 0.1, scale 0.5, horizontal
flip 0.5 and mosaic 1.0; mixup, rotation, shear, perspective and vertical flip
were zero. Batch size 4 was chosen to fit the free Google Colab Tesla T4
environment. No class weighting or resampling was used. The reported protocol used only train and test partitions. Training validation
was disabled (\texttt{val=false}), so there was no validation-based checkpoint
selection or early-stopping criterion. The trained checkpoint was evaluated
after training on the 491-image test partition. The YOLOv8s comparison used
the same dataset files, input size, batch size and 50-epoch setting; its saved
run used seed 42, while the YOLOv8n run used seed 0. These are single-run
comparisons rather than repeated-seed estimates.

\begin{table}[H]
\centering
\caption{Principal training configuration.}
\label{tab:config}
\footnotesize
\setlength{\tabcolsep}{3.5pt}
\begin{tabular}{ll}
\toprule
\textbf{Parameter} & \textbf{Setting}\\
\midrule
Models & YOLOv8n (primary), YOLOv8s (comparison)\\
Input resolution & $640\times640$\\
Classes & Healthy, Unhealthy\\
Epochs & 50\\
Batch size & 4\\
Train / test images & 1,982 / 491\\
Primary hardware & NVIDIA Tesla T4\\
\bottomrule
\end{tabular}
\end{table}

\subsection{Training and Evaluation Algorithm}
Algorithm~\ref{alg:pipeline} summarizes the reported protocol.

\begin{algorithm}[H]
\caption{Banana Surface-Condition Detection Pipeline}
\label{alg:pipeline}
\footnotesize
\begin{algorithmic}[1]
\REQUIRE Banana images $\mathcal{I}$ and YOLO annotations $\mathcal{Y}$
\ENSURE Trained detector $\mathcal{M}^{*}$ and test metrics
\STATE Screen images and verify annotations
\STATE Use the supplied 80\% train and 20\% test partition
\STATE Initialize pretrained YOLO model $\mathcal{M}$
\STATE Set image size $640\times640$, batch size $4$, epochs $E=50$
\FOR{$e=1$ to $E$}
  \STATE Train $\mathcal{M}$ on mini-batches from $\mathcal{D}_{train}$
  \STATE Compute box, classification and DFL losses
  \STATE Update parameters by backpropagation
\ENDFOR
\STATE Set $\mathcal{M}^{*}$ to the final trained checkpoint
\STATE Evaluate $\mathcal{M}^{*}$ on $\mathcal{D}_{test}$
\STATE Generate metrics, curves, confusion matrix and predictions
\RETURN $\mathcal{M}^{*}$ and evaluation outputs
\end{algorithmic}
\end{algorithm}

\subsection{Evaluation Metrics}
Evaluation followed standard object-detection practice, with IoU/AP-style
evaluation consistent with the COCO protocol \cite{lin2014coco}. Precision,
recall and F1-score were
\begin{equation}
P=\frac{TP}{TP+FP},\qquad R=\frac{TP}{TP+FN},
\label{eq:pr}
\end{equation}
\begin{equation}
F1=\frac{2PR}{P+R}.
\label{eq:f1}
\end{equation}
The reported overall F1 of 77.67\% is the harmonic mean of the overall mean
precision and recall. It is not the arithmetic mean of the two class-wise F1
values, which is approximately 77.07\%.

Localization quality was measured using
\begin{equation}
IoU=\frac{|B_p\cap B_g|}{|B_p\cup B_g|},
\label{eq:iou}
\end{equation}
where $B_p$ and $B_g$ are predicted and ground-truth boxes. For $C=2$ classes,
\begin{equation}
mAP_{\tau}=\frac{1}{C}\sum_{c=1}^{C}AP_c(\tau),
\label{eq:map}
\end{equation}
and the stricter metric averaged ten IoU thresholds:
\begin{equation}
mAP_{0.50:0.95}=\frac{1}{10}\sum_{k=0}^{9}mAP_{0.50+0.05k}.
\label{eq:map5095}
\end{equation}
The saved evaluation configuration used an NMS IoU setting of 0.70. Training
losses, PR/F1 curves, confusion matrices and qualitative predictions were also
examined. Runtime was reported as preprocessing, inference and postprocessing
time per image.

\section{Results and Discussion}
\label{sec:results}

\subsection{Overall and Class-Wise Performance}
Table~\ref{tab:results} summarizes the test results. YOLOv8n achieved 81.36\%
precision, 74.30\% recall, 77.67\% F1, 87.04\% mAP@0.50 and 59.33\%
mAP@0.50:0.95. Under the same data files and 640/batch-4/50-epoch setting,
YOLOv8s reached 77.04\% precision, 77.40\% recall, 77.22\% F1, 86.39\%
mAP@0.50 and 59.22\% mAP@0.50:0.95. Thus, YOLOv8s improved recall by 3.11
points but reduced precision by 4.32 points. YOLOv8n remained slightly higher
in F1 and both mAP measures.

\begin{table}[H]
\centering
\caption{YOLOv8 model and class-wise test performance (\%).}
\label{tab:results}
\scriptsize
\setlength{\tabcolsep}{2.5pt}
\begin{tabular}{lccccc}
\toprule
\textbf{Subset/Class} & \textbf{P} & \textbf{R} & \textbf{F1} & \textbf{AP50} & \textbf{AP50--95}\\
\midrule
YOLOv8n test & 81.36 & 74.30 & 77.67 & 87.04 & 59.33\\
YOLOv8s test & 77.04 & 77.40 & 77.22 & 86.39 & 59.22\\
v8n Healthy & 88.81 & 68.00 & 77.02 & 89.14 & 59.24\\
v8n Unhealthy & 73.92 & 80.60 & 77.11 & 84.93 & 59.42\\
\bottomrule
\end{tabular}
\end{table}

Healthy bananas showed higher precision but lower recall, whereas unhealthy
bananas showed the opposite pattern. The training partition contained 2,355
healthy and 2,027 unhealthy instances. No class weighting or resampling was
used, so the moderate frequency imbalance may contribute to this asymmetry;
it should not be attributed only to detector behavior. Similar fruit-defect
studies report difficulty when abnormal texture or coloration overlaps with
normal skin appearance \cite{feng2023orange}.

The YOLOv8n checkpoint was 5.96 MB, while the compared YOLOv8s checkpoint was
21.48 MB. Since YOLOv8s did not improve F1 or mAP in this run, YOLOv8n provided
the better checkpoint-size/performance trade-off. Because only one run per
model was available and the saved seeds differed, the small metric differences
should be treated as descriptive rather than as confidence-tested model
rankings.

\subsection{Training and Diagnostic Analysis}
\begin{figure}[!b]
\centering
\includegraphics[width=\columnwidth]{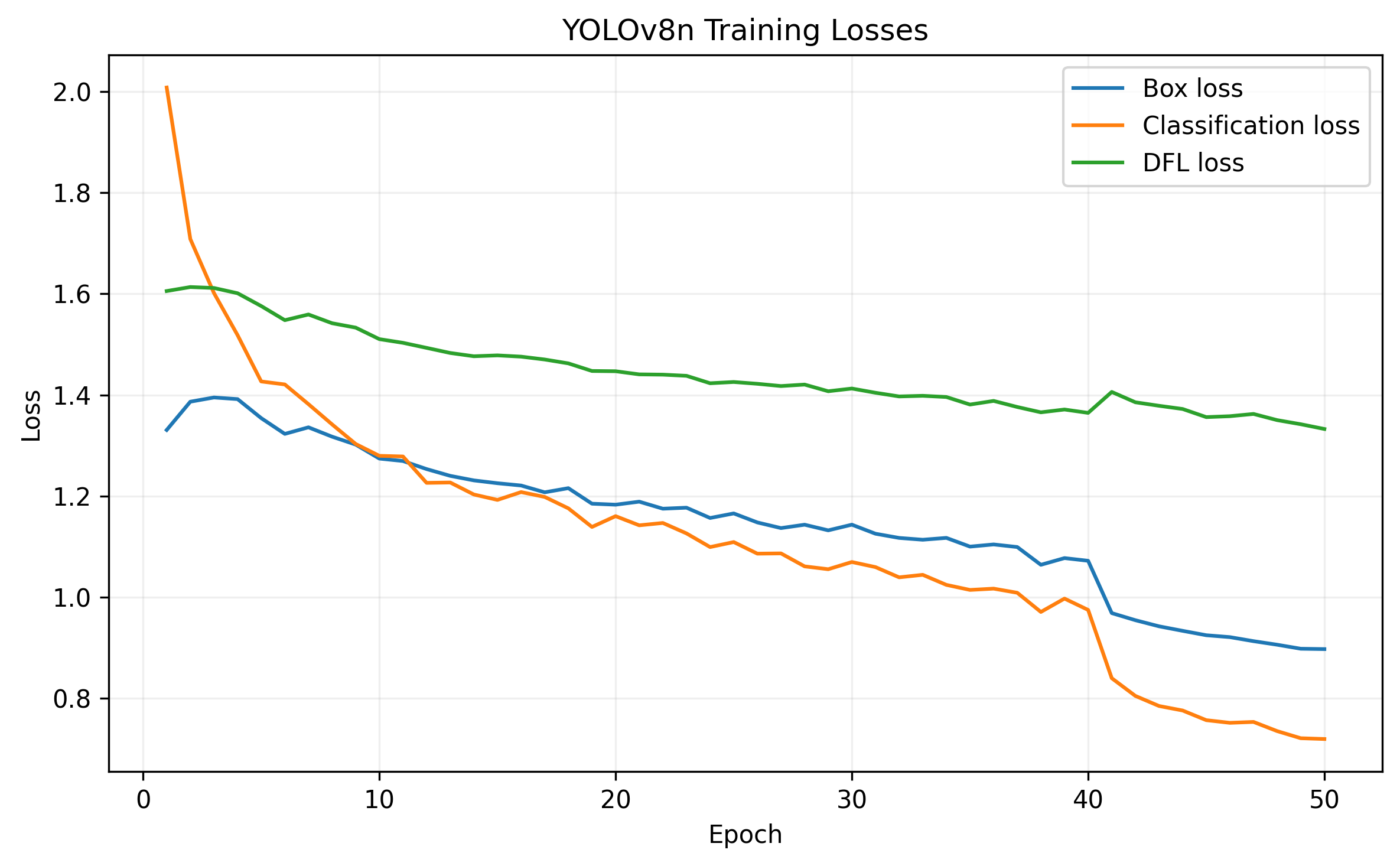}
\caption{Training-loss behavior of the selected YOLOv8n model.}
\label{fig:loss}
\end{figure}

The box, classification and DFL components decreased and stabilized during
training (Fig.~\ref{fig:loss}), showing convergence of the optimization
objective. Since no validation partition was used, training-loss convergence
alone is not evidence of generalization.

\begin{figure}[H]
\centering
\includegraphics[width=\columnwidth]{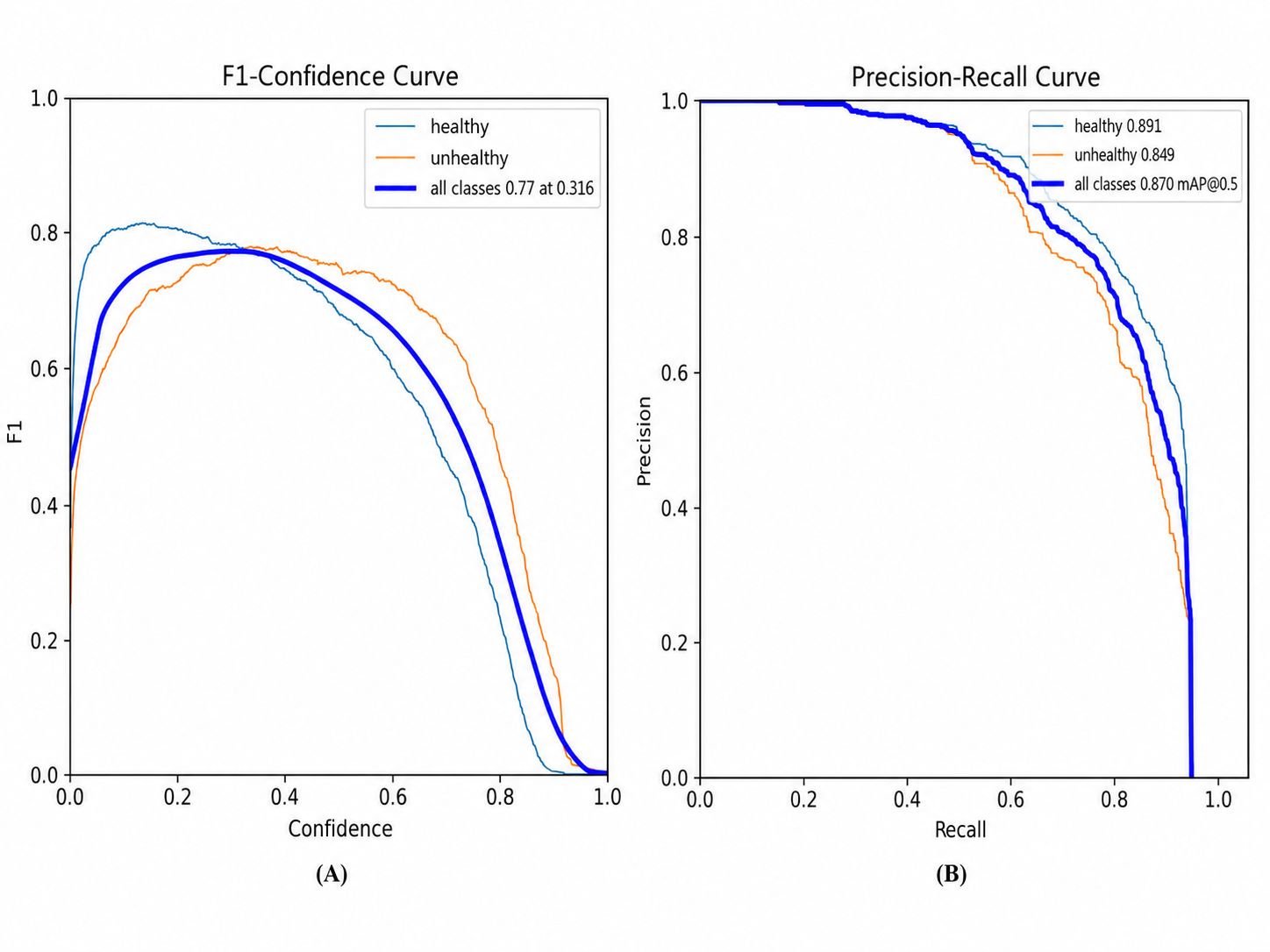}
\caption{Combined precision--recall and F1-confidence curves on the test partition.}
\label{fig:diagcurves}
\end{figure}

The precision-recall curve gives AP@0.50 values of 0.891 for healthy and
0.849 for unhealthy instances, with all-class mAP@0.50 of 0.870. The
F1-confidence curve reaches approximately 0.77 at confidence 0.316
(Fig.~\ref{fig:diagcurves}). This value was obtained from the test curve and is
reported only as a descriptive diagnostic; it was not selected on a validation
set and should not be treated as a validated deployment threshold.

The 27.71-point gap between mAP@0.50 (87.04\%) and mAP@0.50:0.95 (59.33\%)
shows that localization becomes less consistent as the required overlap
becomes stricter. Overlapping bananas, partial occlusion and uncertainty in
manual whole-fruit box boundaries are plausible contributors, but these causes
were not isolated experimentally.

\begin{figure}[H]
\centering
\includegraphics[width=\columnwidth]{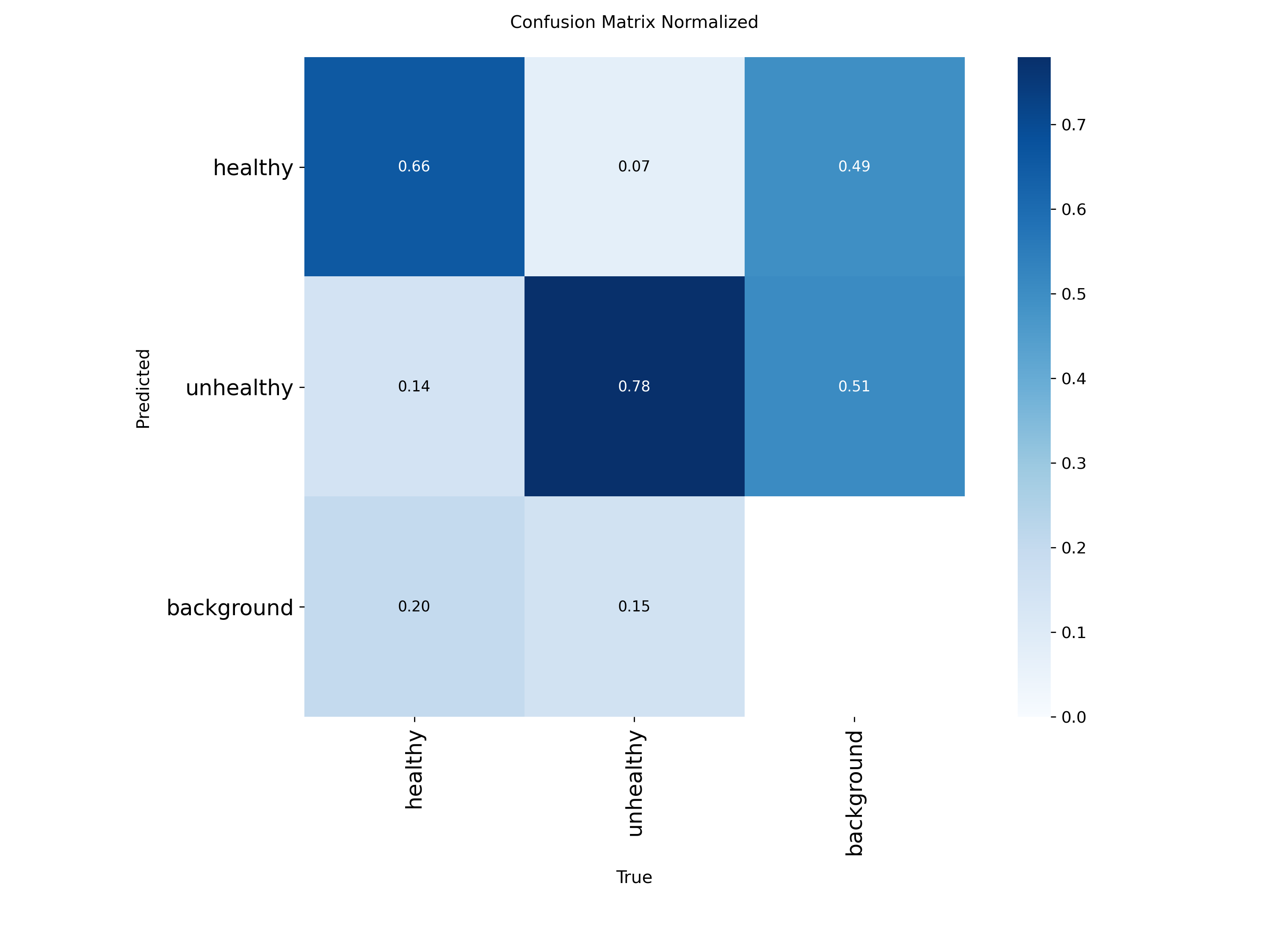}
\caption{Normalized confusion matrix for the test partition.}
\label{fig:cm}
\end{figure}

The normalized confusion matrix (Fig.~\ref{fig:cm}) showed asymmetric
cross-class confusion: 14.24\% of true healthy instances were predicted as
unhealthy, whereas 7.05\% of true unhealthy instances were predicted as
healthy. Natural discoloration, shadows and subtle scars are plausible
contributors, but these factors were not isolated experimentally
\cite{wu2024dneyolo}.

\subsection{Qualitative Results and Efficiency}
\begin{figure}[H]
\centering
\includegraphics[width=\columnwidth]{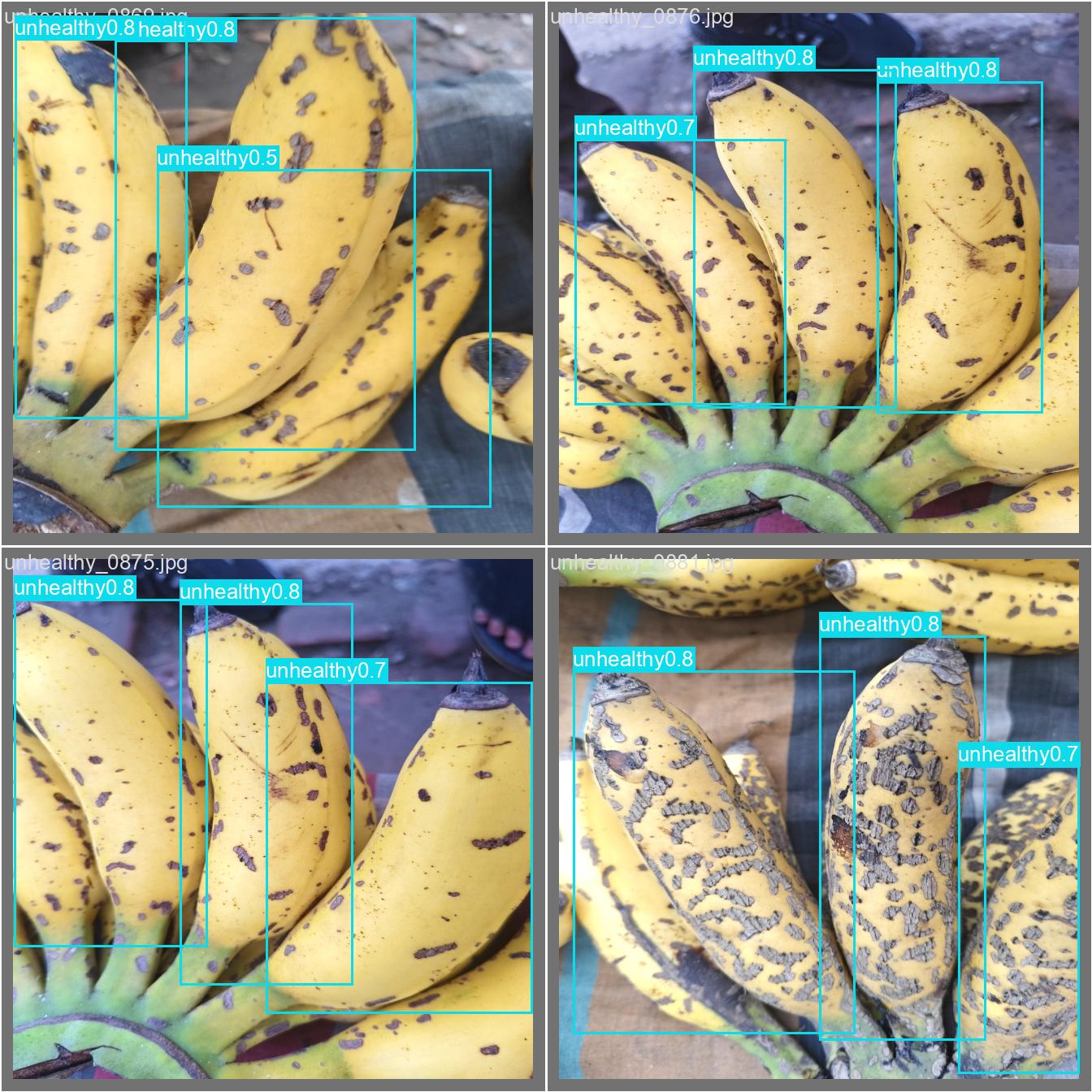}
\caption{Representative test predictions with instance-level labels.}
\label{fig:qual}
\end{figure}

Representative predictions in Fig.~\ref{fig:qual} show that the detector can
localize multiple bananas within one frame and assign separate condition
labels. This instance-level behavior is useful when healthy and visibly
abnormal bananas appear together.

The YOLOv8n model contained 3,006,038 parameters and occupied 5.96 MB. Average
preprocessing, inference and postprocessing times were 1.55, 6.54 and
1.60 ms/image, respectively. The inference-only value corresponds to
approximately 153 FPS on the Tesla T4. YOLOv8s timing was not measured under
the same reporting procedure, so runtime is not compared directly. The
YOLOv8n result nevertheless supports a compact deployment footprint
\cite{karacaoglu2025kiwi}.

\subsection{Limitations}
The study used a single 80/20 split and one run per model; no separate
validation set, repeated-seed analysis, cross-validation or confidence
intervals were available. A later exact-hash audit of the released split found
repeated image content across train and test, which can make the reported test
score optimistic and prevents a strong claim of group-level independence.
The dataset was collected from four Bangladeshi market locations, the cultivar
identity was not recorded and no external dataset was tested. The two broad
labels describe visible appearance rather than biological cause, and no
inter-annotator agreement statistic is available. These limitations motivate a
future group-aware split, repeated runs, broader acquisition conditions and
external evaluation.

\section{Conclusion}
This study evaluated lightweight YOLOv8 models for instance-level banana
surface-condition detection using 2,473 images and 5,827 annotated instances.
On the 491-image test partition, YOLOv8n achieved 81.36\% precision, 74.30\%
recall, 77.67\% F1-score, 87.04\% mAP@0.50 and 59.33\% mAP@0.50:0.95.
YOLOv8s increased recall but did not improve F1 or mAP and used a much larger
checkpoint. The 0.316 confidence value is a test-derived diagnostic rather
than a validated operating threshold. With 3.01 million parameters, a
5.96~MB checkpoint and 6.54~ms/image measured inference time on the Tesla T4,
YOLOv8n remains a useful compact baseline for this application. The conclusions
are limited to the supplied split and local acquisition conditions; the system
does not perform biological diagnosis or identify the cause of surface damage.

\section{Data and Code Availability}
\label{sec:data_availability}

The dataset, YOLO annotations, experiment configuration, checkpoints, audit
records and supporting code are available at
\url{https://github.com/mehedinaeem/Bitol_Computer_Vision_System}.
The repository documents the 1,982/491 train/test organization, the
640-pixel training configuration, checkpoint
provenance and dataset-hash audit. These records are provided so that the
reported split, known data-quality limitations and model outputs can be
examined directly.

\bibliographystyle{IEEEtran}
\bibliography{biblio}

\end{document}